\documentclass[letterpaper, 10pt, conference]{ieeeconf} 
\IEEEoverridecommandlockouts                              

\usepackage[vlined, ruled, linesnumbered, commentsnumbered]{algorithm2e}
\usepackage{amsmath}
\usepackage{cite}
\usepackage{color}
\usepackage{xcolor}
\definecolor{chred}{rgb}{0.8,0,0}
\definecolor{chgrey}{rgb}{0.5,0.5,0.5}
\usepackage{graphicx}
\usepackage{subcaption}
\usepackage{dblfloatfix}
\usepackage{eqlist}
\usepackage{txfonts}
\usepackage{url}
\usepackage{footmisc}
\usepackage{booktabs}
\usepackage{balance}\usepackage{flushend}

\usepackage{multirow}
\usepackage[para,online,flushleft]{threeparttable}
\usepackage[final]{pdfpages}

\usepackage{threeparttable}

\title{\LARGE \bf
Combined Task and Motion Planning for a Dual-arm Robot to \\Use a Suction Cup Tool}

\author{Hao Chen$^{1}$, Weiwei Wan$^{1,2,*}$, and Kensuke Harada$^{1,2}$
\thanks{
$^{1}${Graduate School of Engineering Science, Osaka University, Japan.}
$^{2}${National Inst. of AIST, Japan.} *{Correspondent author: Weiwei Wan,}
{\tt\small wan@sys.es.osaka-u.ac.jp}
}}

\begin{document}

\maketitle
\thispagestyle{empty}
\pagestyle{empty}

\begin{abstract}

This paper proposes a combined task and motion planner for a dual-arm robot to use
a suction cup tool. The planner consists of three sub-planners --
A suction pose sub-planner and two regrasp and motion sub-planners.
The suction pose sub-planner finds all the available poses for a suction
cup tool to suck on the object, using the models of the tool and the object.
The regrasp and motion sub-planner builds the regrasp graph that represents
all possible grasp sequences to reorient and move the suction cup tool from an initial pose to a goal pose. 
Two regrasp graphs are used to plan for a single suction cup 
and the complex of the suction cup and an object respectively.
The output of the proposed planner is a sequence of robot motion that uses a suction cup tool
to manipulate objects following human instructions.
The planner is examined and analyzed by both simulation experiments and real-world executions using several real-world tasks.
The results show that the planner is efficient, robust, and can generate sequential transit and transfer robot motion
to finish complicated combined task and motion planning tasks in a few seconds.

\end{abstract}

\section{Introduction}

\begin{figure}[t]
    \centerline{\includegraphics[width=.9\linewidth]{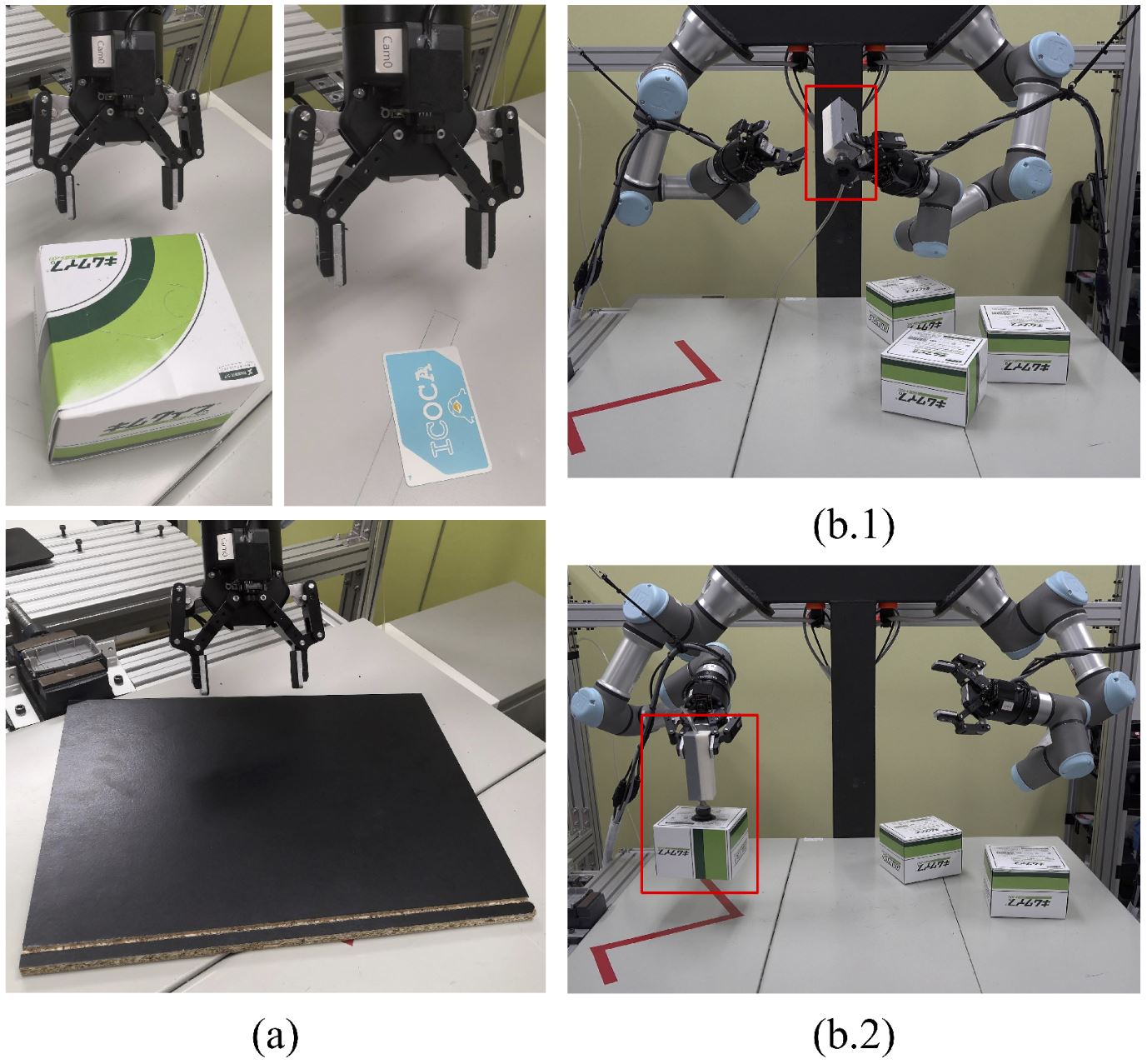}}
    \caption{(a) A general two-finger robot hand that does not have enough stroke to grasp
    objects that are too large or too thin. (b.1-2) Using a suction cup tool to pick up 
    and move an object that is larger than the stroke of the robotic gripper.}
    \label{hand}
\end{figure}

Robots are widely used to pick up and move various types of objects in manufacturing. 
When objects are too large or too thin for a general two-finger robotic gripper to 
catch and move, like Fig.\ref{hand}(a). A common solution is to develop 
specialized grippers \cite{bibhasegawa2017three,takahashi2013flexible,yamaguchi2013development}. 
However, there are some drawbacks like the high cost in developing a special gripper, 
the need of special equipment like tool changers to switch between the different robotic grippers, 
the annoying grasping design for each gripper, etc.
These drawbacks could be solved by developing an intelligent robot that could use
tools like humans. 
For example, Hu et al.~\cite{hu2019designing} designed low-cost
tools for the robot with two-finger parallel grippers to pick different types of objects.
Each tool is inexpensive and the tools can be switched by pick and place.
The tools reduce the cost
and extend the flexibility of a traditional robot.
Hu et al.'s paper focused on the mechanical design and stability analysis of a gripping tool and
developed a robotic system that can use various variations of the gripping tool to perform difficult tasks.
This work extends Hu et al.'s study. Instead of focusing on design and stability, 
this work discusses the motion planning algorithms to use
the tools, with a special focus on the suction cup tool.

Using a suction cup tool involves two steps: The first step is to reorient a single suction 
cup tool from its initial pose to the object to be picked.
The second step is to pick up the object, reorient the complex of the suction cup tool,
and move it to the goal. Thus, planning the motion to use a suction cup tool
needs two independent manipulations. By using the (i) the pose of the tool, 
(ii) the pose of the object to be moved, (iii) the goal pose of the object,
and (iv) the model of the suction cup tool and the object, this paper proposes a 
combined task and motion planner which can automatically find the motions
for a dual-arm robot to use a suction cup tool to pick up objects and pile them up following
human instructions.

The combined task and motion planner consists of a suction pose sub-planner
and two regrasp and motion sub-planners. The suction pose sub-planner finds 
all the available poses for a suction cup tool to suck on the object, using the
models of the tool and the object. The regrasp and motion sub-planner builds and searches the
regrasp graph that represents all possible grasp sequences to reorient and
move the suction cup tool from an initial pose to a goal pose.
The first regrasp and motion sub-planner solves the manipulation of
a single suction cup tool. The second regrasp and motion sub-planner solves the manipulation of
the tool-object complex. They
search the regrasp graphs to find a sequence of grasps and plan the path
between adjacent grasps in the sequence using a sample-based motion planning algorithms.

The contribution of this paper is the development of the combined planner that
enables a dual-arm robot to use the suction cup tool to perform pick-and-place tasks.
All details related to the robot, including the suction poses, the handovers, 
the motions and the pick-and-place order are determined automatically by the planner
considering the geometric and kinematic constraints.

The planner is examined and analyzed by
both simulation experiments and real-world executions using several real-world tasks.
The results showed that the planner is efficient and robust. 
The planner can successfully plan the feasible and collision-free path to move the objects
to the given place by using the suction cup tool within a few seconds.

The paper is organized as follows. Related work is discussed in Section II. 
An overview of the proposed planner is presented in section III.
The implementation details are presented in Section IV.
Experiments and analysis are presented in Section V.
Conclusions are drawn in Section VI.

\section{Related Work}

This work proposes a combined task and motion planner for
a dual-arm robot to use a suction cup tool. Thus, we review
the related literature in (i) the combined task and motion planning, and
(ii) the dual-arm manipulation planning.

\subsection{Combined task and motion planning}
The combined task and motion planning plans the task order at the high level and
considers the geometric constraints and performs the motion planning for the task at the
low level. Previously, many different kinds of combined task and motion planners were proposed,
like the a Symbolic Move3D (aSyMov)~\cite{cambon2009hybrid}, the Hierarchical Task 
Networks for Mobile Robot (HTN-MM)~\cite{wolfe2010combined}, 
the Hierarchical Planning in the Now (HPN)~\cite{kaelbling2010hierarchical}\cite{faser2016}.
More recently, Bidot et al.~\cite{bidot2017geometric} structured a geometric backtracking
task and motion planner for a humanoid robot to move the cups to the tray, 
which reconsiders past geometric decisions so as to fulfill the preconditions of the present action. 
Su{\'a}rez et al.~\cite{suarez2018interleaving} presented a framework combining
the hierarchical task planning, motion planner, and geometric reasoning to plan the 
motion of a bimanual task using a dual-arm robot. The framework divides the total task into small
sub-goals and checks the geometric constraints and collisions at the sub-goal level.
Colledanchise et al.~\cite{colledanchise2019} developed a blended reactive planning and acting framework using BT (Behavior Trees).

\subsection{Dual-arm manipulation planning}
Plenty of studies concentrate on dual-arm manipulation planning. 
For example, Harada et al.~\cite{harada2014manipulation} presented regrasp 
and handover planning of dual-arm robots by considering the multi-modal 
configuration spaces of the objects. Su{\'a}rez et al.~\cite{suarez2015using}
proposed a planning method for the dual-arm robot by sampling the behaviors of human bi-manual tasks to
find a configuration space and search the path in this configuration space. 
More recently, Moriyama et al.~\cite{moriyama2019dual} presented the dual-arm assembly planning
considering the gravitational constraints. The blended planner proposed by Colledanchise et al.
could also plan dual-arm handover motion.

Compared with the above studies, we proposed a combined task and motion planner 
that considers the case of using a suction cup tool. The proposed planner automatically decides the suction
pose on the object, the motion to move the suction cup tool to the suction pose,
as well as the motion to move the complex of tool and object to the indicated position.
The planner uses two regrasp graphs in the task level for determining the grasp sequence
and use the sampling-based algorithm in the motion level to plan the path in the low level.
To our best knowledge, this study is the first combined task and motion planner that plans motion
for a dual-arm robot to use tools.

\section{Overview of the Proposed Planner}
\begin{figure*}[!htbp]
    \centerline{\includegraphics[width=.97\textwidth]{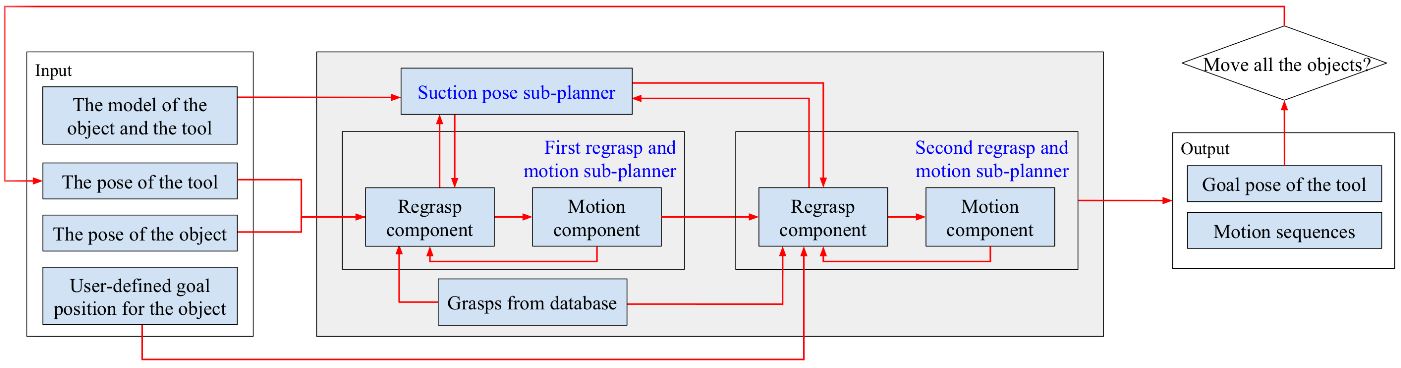}}
    \caption{The workflow of the proposed planner. The shadow area is the 
    main body. It includes a suction pose sub-planner and
    two regrasp and motion sub-planners.}
    \label{flowchart}
\end{figure*}

This section presents an overview of the proposed planner. The workflow is shown in Fig.\ref{flowchart}.

In the beginning, the suction pose sub-planner receives the model of an object and a suction cup tool as the input.
It finds all the feasible suction poses on the object for the suction cup tool using
the models of the object and the tool.
The planned suction poses are used by both the 
first and second regrasp and motion sub-planners. They check the IK-feasible and collision-free grasp configurations
of the suction cup tool for initial and goal object poses simultaneously.
If the grasps of the tool exist at both the initial and goal poses of the object, the suction pose
will be chosen. Otherwise, the regrasp and motion sub-planners will iterate to another suction pose
until a feasible one is found.

The first regrasp and motion sub-planner receives (i) the pose of objects, 
(ii) the pose of a suction cup tool, and (iii) the suction pose on the object, as the input.
The regrasp component of the sub-planner finds a sequence of collision-free
and IK-feasible grasps to reorient
and move the tool from the initial pose to the suction pose by building and searching a regrasp graph
using the input.
The motion component of the sub-planner plans the motion between
every two adjacent grasps in the sequence obtained by the 
regrasp component. The results of the first regrasp and motion sub-planner include
(i) the grasp pose for the tool and (ii) the motion sequences to move the tool. 

The second regrasp and motion sub-planner receives some user-defined goal poses of the object as the input. 
The fundamental algorithm of the second one is similar to the first one. 
The regrasp component of the sub-planner builds the
regrasp graph according to the initial grasp, the handover grasps, and the grasps of the tool at the goal pose
of the object. It searches the regrasp graph and finds a sequence of grasps.
The motion component plans the motion between each adjacent grasp pair in the sequence.
The output of the second regrasp and motion
sub-planner includes (i) the goal pose of the tool, (ii) the final grasp pose of the robot,
and (iii) the motion to move the complex of the tool and the object to the instructed pose. 

In cases where the motion components in the two regrasp and motion sub-planners 
cannot find the path between an adjacent grasp pair, the planner backtracks to the regrasp component
and searches another sequence of the grasps. In cases where no sequence is found by the regrasp components,
the planner backtracks to the suction pose sub-planner, reconsiders another
suction pose on the object, and plans again. After moving the object to the goal position, 
the system will check the remaining objects. If there are remaining objects,
the final grasp pose of the tool will be sent to the regrasp component of the first regrasp and motion
sub-planner as the initial grasp for combined planning. The object that is closest to the given goal
place will be considered first.

The planner is purely geometric. We assume that the suction cup tool can
stably suck up the object. We do not consider the problems
that the object is too heavy to be picked up or the tool slips or drops during moving.

\section{Details of the Sub-planners}
This section presents the details of the suction pose sub-planner and the two regrasp and motion sub-planners.

\subsection{Suction pose sub-planner}
\begin{figure}[!htbp]
    \centerline{\includegraphics[width=\linewidth]{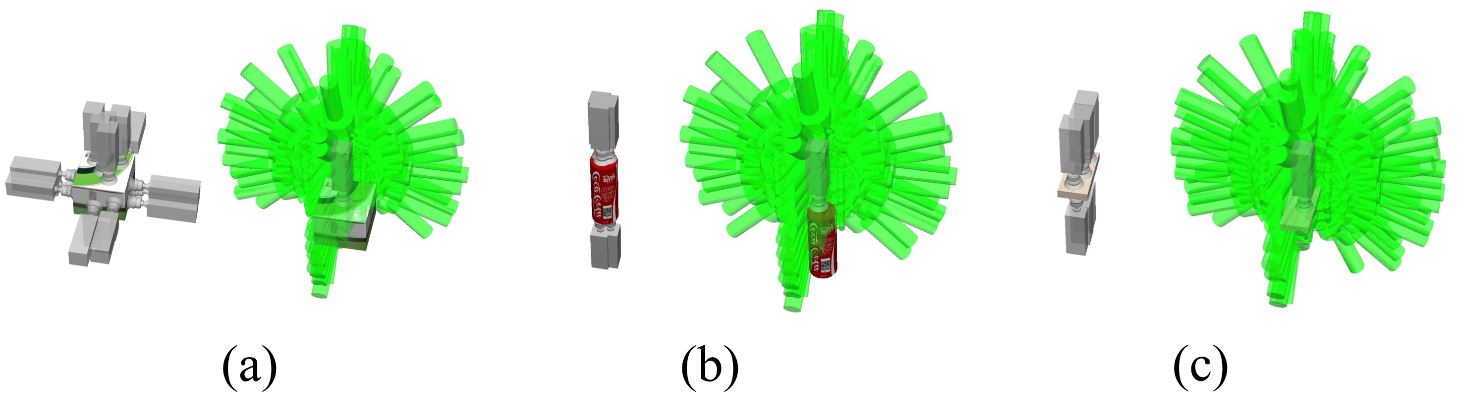}}
    \caption{(a) The suction poses for a tissue box and the possible grasp 
    poses of the suction cup tool on one suction pose.
    (b) The suction poses for a Coke can and the possible grasp 
    poses of the tool. (c) The suction poses for a Domino blocks and the possible grasp. }
    \label{suctionpose}
\end{figure}

We use the method proposed in Wan et al. \cite{wan2016achieving} to calculate the precise suction poses
on the object. The suction poses are essentially the relative transformation between a suction cup tool and an object
according to the models of the tool and the object. The method is able to find
the grasp poses for the suction cups and the parallel grippers. The workflow of the method is (i) find the planar facets,
(ii) sample the facets, and (iii) find the candidate samples for attaching a suction cup tool. Fig.\ref{suctionpose}
shows the planned suction poses for various objects. By using the model of a suction cup tool
and an object, the suction pose sub-planner computes the relative transformation set 
$S_{transfrom} = \{ \boldsymbol{T}^t_1, \boldsymbol{T}^t_2,... \boldsymbol{T}^t_n\}$ of the tool on the object.
The relative transformation is $\boldsymbol{T}^t=\{p^t, R^t\}$, where the $p^t$ and $R^t$ respectively
represents the position and orientation of the suction cup tool relative to the object.
The pose $\boldsymbol{P}^{tool}=\{p^{tool}, R^{tool}\}$ of the suction cup tool on the suction pose can be denoted as
\begin{equation}
p^{pose} = R^t p^{obj} + p^t
\end{equation}
\begin{equation}
R^{tool} = R^{t} R^{obj},
\end{equation}
where $p^{obj}$ and $R^{obj}$ are the position and orientation of the object, respectively.

\subsection{The first regrasp and motion sub-planner}

The regrasp and motion sub-planner includes two components -- a regrasp component and
a motion component.

The regrasp component is based on the framework proposed in \cite{wan2019preparatory}. 
The input of the regrasp component is (i) the initial pose of the tool, (ii) the relative suction pose $\boldsymbol{T}^t$,
(iii) the goal pose of the object, and (iv) the specific grasp of the tool. The fourth input is optional. 
 It is used in the multiple objects situation so that the final grasp
of the tool in the last round could be specified as the initial grasp in the current round. 
The output of the first regrasp and motion sub-planner is (i) the goal pose of the tool, 
(ii) the grasp of the tool, and (iii) the motions to move the tool to the suction pose on the object. 
The regrasp component first loads the grasp poses and handover grasp pairs of the suction cup tool from the database (see details in \cite{wan2019preparatory}). 
The grasp poses are the possible grasps for the suction cup tool. The handover grasp pairs are the IK-feasible 
and collision-free grasps of the left arm and right arm of the robot at an optimal handover space in front of the robot, 
where both arms have the maximum manipulability. These grasps are calculated and stored in the database
in advance to reduce the time of building the regrasp graph. The method of calculating these grasps
is the same as the way to calculate the suction poses on the object. After loading the grasps from the database,
the regrasp component calculates
the IK-feasible and collision-free grasp configurations in the initial pose of the tool and in the suction
pose of the object. In the case that is no grasps on the suction pose of the object at the start
pose, the regrasp component backtracks to the suction pose sub-planner and tries another suction pose
until both first and second regrasp components can find grasp configurations for
the tool on the suction poses of the object at the start pose and goal pose.

\begin{figure}[htbp]
    \centerline{\includegraphics[width=.8\linewidth]{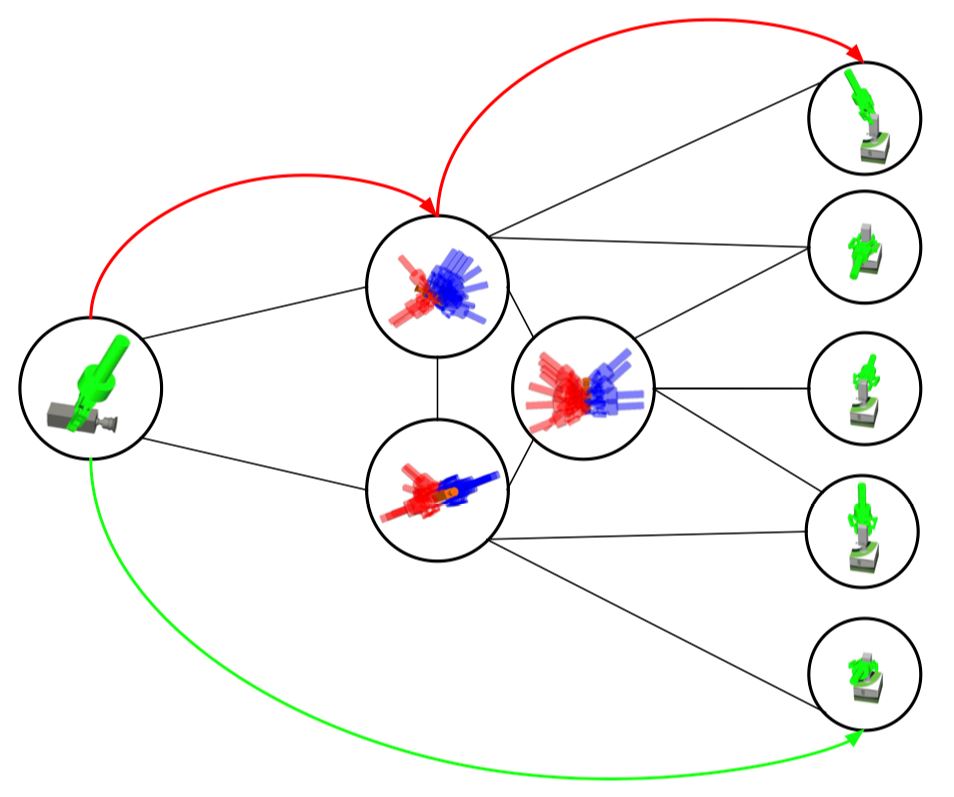}}
    \caption{An example of a regrasp graph.}
    \label{regrasp}
\end{figure}

An example of the regrasp graph is shown in Fig.\ref{regrasp}. A regrasp graph is
constructed depending on the initial grasps, goal grasps, and some intermediate
handover grasps. The node in the left part of the regrasp graph represents all the IK-feasible
and collision-free grasps for the tool at the initial position. The nodes in the middle layer encode
several possible handover grasps. The nodes in the right layer represent all grasps of the tool for
the chosen suction pose on the object. Each node encodes a grasp of the suction cup tool.
The nodes are connected by the edge (black line). The regrasp component recursively searches through edges
to find a sequence of grasps to reorient and move the single tool from its initial pose to a suction pose
on the object. The red line in the upper part of Fig.\ref{regrasp} shows one example of the sequence search
result between the initial and goal grasp. It means that the single robot arm cannot move
the tool from the initial pose to the suction pose. A handover is needed to reorient and use the other hand to move the tool to
the suction pose. The green line at the bottom of the demonstration is another example of
the sequence search result, which means that one robot arm can move the tool from the
initial pose to the suction pose directly.
Instead of adding all the grasps of the tool on every suction poses into the regrasp graph,
we just add one set of grasps of the tool for one chosen suction pose every time.
Because we have a large number of common grasps and handover grasps, adding too many grasps of the tool into
the regrasp graph makes the graph complicated and redundant. It is thus advisable to
add only a few grasps to decrease the search time.

The motion component plans the motion between every adjacent two grasps of the grasp sequence 
by using the RRT-connect algorithm \cite{kuffner2000rrt}.
RRT-Connect traverses a configuration space by generating two random search trees from
both the start and the goal. The two trees randomly sample a direction in configuration space and append
a node towards the random direction with the shortest distance to the tree. The newly appended
point will be the end node of the tree. The path is found when the distance between 
the end nodes of the two trees is smaller than a threshold. 
The algorithm does not necessarily find a path. When there is no path found, 
the edge between these two grasps in the regrasp graph is deleted.
The motion component backtracks to the regrasp component and re-searches the regrasp 
graph to find a new grasp sequence and plan the motion again.
If all edges are deleted and no results are planned,
the planner will remove the node and the edge of the grasps
in the old suction pose and rebuild the edge and the node in a new suction
pose. The object is treated as a stationary obstacle in motion planning.

\subsection{The second regrasp and motion sub-planner}
The second regrasp and motion sub-planner is very similar to the first one. 
The input of the second planner is (i) the final grasp in the first regrasp and motion sub-planner, 
(ii) the goal pose of the object, (iii) the relationship 
$\boldsymbol{T}^t$ between the suction tool cup and the object.
The output is (i) the motion to move the tool and object to the given place,
(ii) the final grasp of the tool, and (iii) the final pose of the tool. 
The goal pose of the tool is computed using the pose of the object and
the relationship $\boldsymbol{T}^t$.
Thus, the problem of moving the object is essentially equal to moving the suction cup tool. 
The grasp configurations of the tool on the suction pose of the object at the goal
pose are calculated and checked. If there are no grasp configurations
at the goal pose of the object, the planner backtracks to the suction pose
sub-planner to choose another suction pose. 

At the bottom, the second regrasp and motion sub-planner
reuses the first regrasp component's regrasp graph.
The initial grasp of the tool is the final grasp of the tool in the first regrasp 
and motion sub-planner. The goal grasp of the tool can be calculated according to 
the given goal pose of the object. The node in the left layer encodes the one 
initial grasp. The nodes in the middle layer encode several possible handover
grasps. The nodes in the right layer represent all grasps of the 
tool for the chosen suction pose on the object. 
The grasp and motion sequence is determined by searching the
regrasp graph and by using the RRT-connect motion planning algorithm.

Note that the second regrasp and motion sub-planner regards
tools and objects as a whole during motion planning. We reuse the handover grasps pair
in the first regrasp component. However, when calculating these handover grasp pairs, 
it only considers the collision between the robot hand and the suction cup tool. 
Detecting the collision between the hand and the object during the handover needs to be further included. 
Some grasps on the suction cup tool are deleted in the second sub-planner
because of the collision between the hand and the object. 


\section{Experiments and Analysis}

\begin{figure}[!htbp]
    \centerline{\includegraphics[width=.85\linewidth]{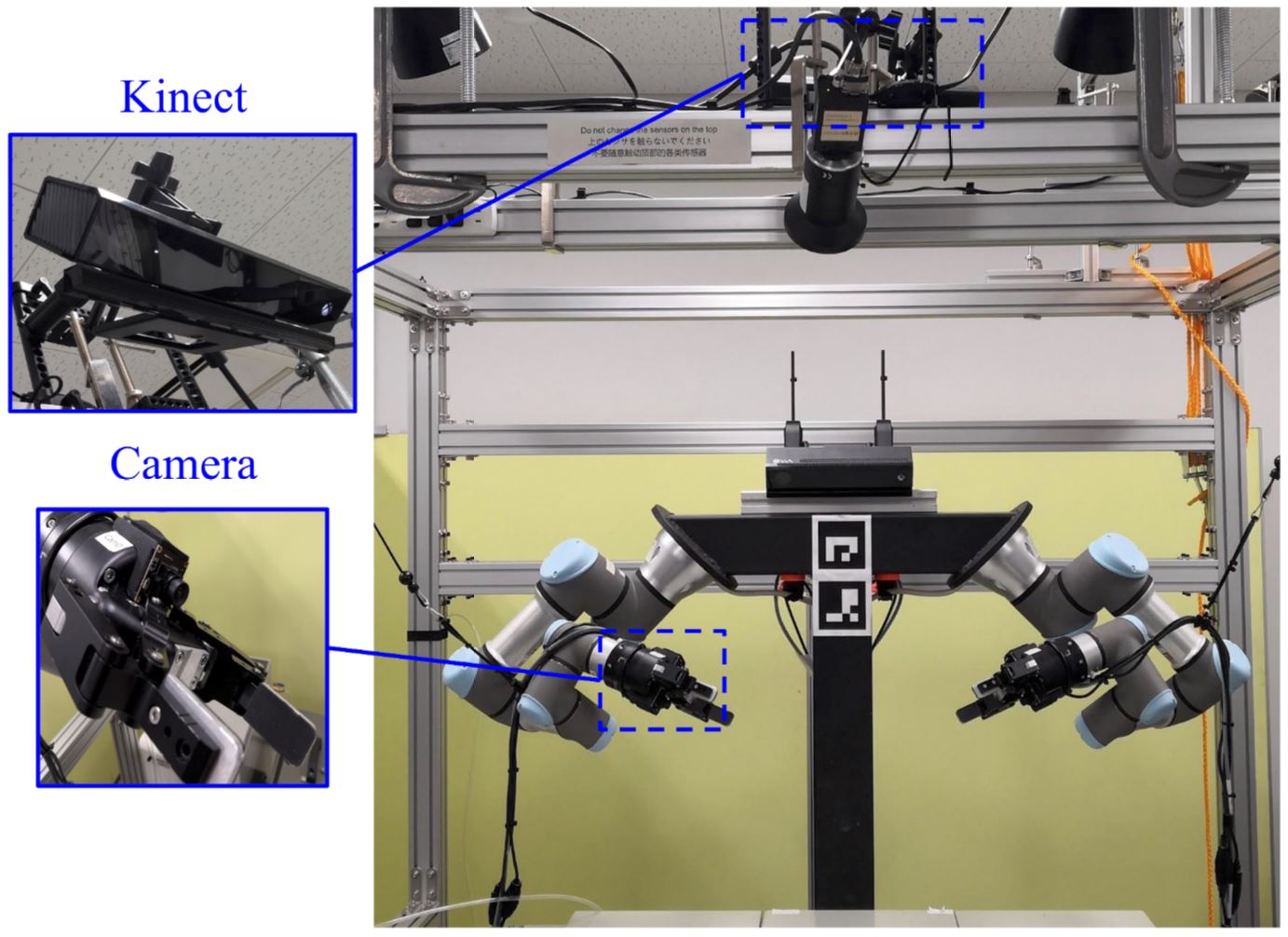}}
    \caption{The experiment environment.}
    \label{envir}
\end{figure}

The proposed planner is validated using a dual-arm UR3 robot in several scenarios.
The experiment environment is shown in Fig.\ref{envir}. Two UR3 robots equipped
with Robotiq F-85 parallel finger grippers are symmetrically installed to a steel body.
There is an RGB camera mounted on the right hand of the robot.
A Kinect is installed on top of the frame to recognize the objects on the table. 

\subsection{Visual detection}
\begin{figure}[htbp]
\centerline{\includegraphics[width = .95\linewidth]{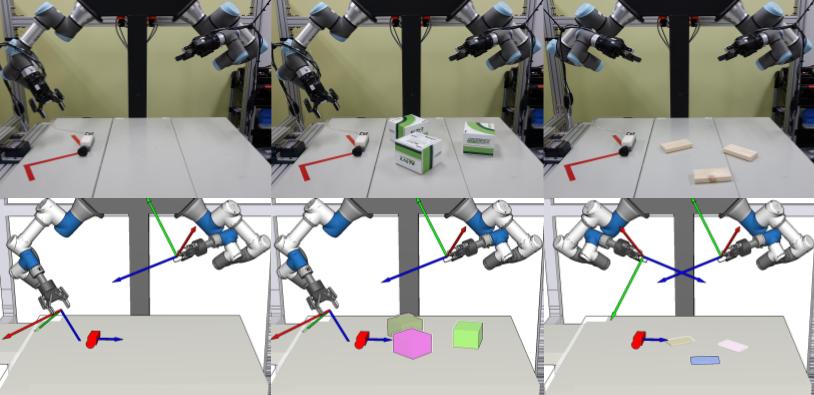}}
\caption{The left column is the result of tool recognition using an AR marker. 
The middle and right columns show the result of recognizing tissue boxes and the Domino blocks using the Kinect
mounted on top of the aluminum frame.}
\label{kinectAndArMarker}
\end{figure}

We build a visual system to capture the position of the tool and objects on the table.
The AR marker attached to the tool is used for detecting the pose of the tool.
A camera mounted on the right hand of the robot is used to find and recognize the marker.
The left column of Fig.\ref{kinectAndArMarker} shows the result of detecting the
position of the suction cup tool by using an AR marker.

In order to make the proposed combined task and motion planner work for different kinds of objects, 
we do not prepare the model of the object to be moved. Instead, we use point cloud
information to detect both the pose and the model of the object. The Kinect mounted on
the top of the frame generates the point cloud information of the objects on the desk.
We assume the model of the object to be moved is a cube or a cylinder. When processing the point cloud information, 
we segment the top facet of the object for suction. The height of the object is inferred
by the distance between the desk and the top facet of the object. The model of the object
is obtained according to the shape of the top facet and the height of the object.
The middle and right columns of Fig.\ref{kinectAndArMarker} show some examples of the recognized objects.
Although there are many feasible suction poses on the top facet of the object,
the central one is chosen as a priority pose for the planner since
it is in most cases the most stable suction pose.

\subsection{Simulation}
\begin{figure*}[!htbp]
\centerline{\includegraphics[width=.97\textwidth]{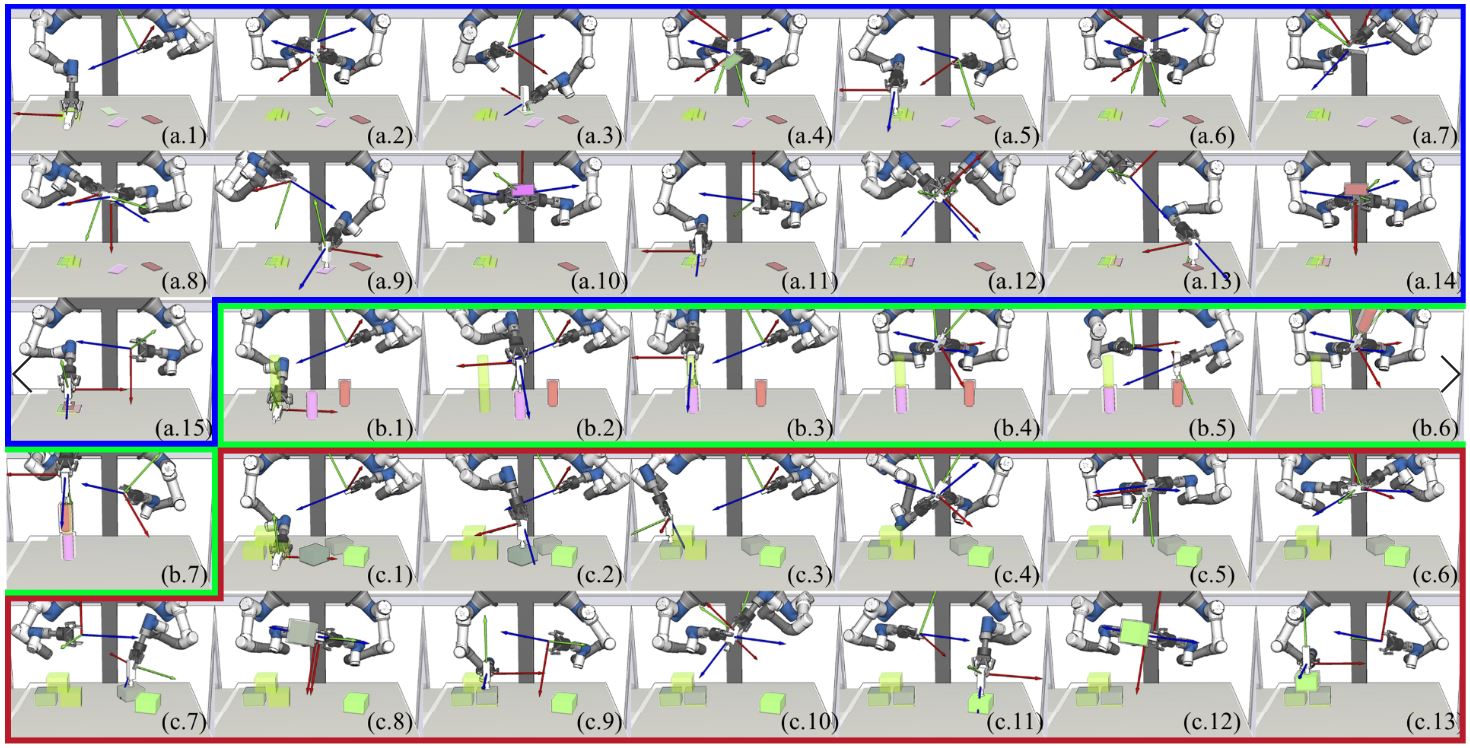}}
\caption{The simulation results of moving three exemplary objects. 
(a) The robot pile up three Domino blocks. 
(b) The robot stacks up two Coke cans. (c) The robot piles up three tissue boxes. }
\label{simulation}
\end{figure*}

In the simulation experiments, we use the proposed planner to plan motion to move three different
kinds of objects from some random initial poses to some human-instructed goal poses. 
The objects used in the experiment include Domino blocks, Coke cans, and tissue boxes. Fig.\ref{simulation}
shows the motion results in the simulation experiments.
The green transparent parts on the table are the goal poses of the objects.
The colored parts on the table are the initial object poses captured and reconstructed using
the vision system. Fig.\ref{simulation}(a.1)-(a.15) pile up three Domino blocks at a specific position.
Fig.\ref{simulation}(b.1)-(b.7) stack up two Coke cans. Fig.\ref{simulation}(c.1)-(c.13)
pile up three tissue boxes.





\begin{table*}[!htbp]
\centering
\renewcommand{\arraystretch}{1.2}
\caption{\label{conf0}Time consuming of the simulation experiments}
\label{tb:timeconsuming}
\resizebox{0.95\linewidth}{!}{%
\begin{threeparttable}
\begin{tabular}{cccccccc}
\toprule
 & & \multirow{
 2}{*}{Suction cup planner} &
 \multicolumn{2}{c}{{First regrasp and motion sub-planner}}
 &
  \multicolumn{2}{c}{{Second regrasp and motion sub-planner}}
  & \multirow{2}{*}{{Total time}}
 \\
 & & & Regrasp component & Motion component & Regrasp component & Motion component &\\
 \midrule
 \midrule
 \multirow{ 4}{*}{Domino block} 
 & 1st object &  0.06 $s$  & 14.67 $s$ & 9.12 $s$ &  7.71 $s$ & 5.38 $s$ & 36.94 $s$\\
 & 2nd object &  0.06 $s$  & 8.01 $s$  &  9.84 $s$ & 7.96 $s$ & 6.29 $s$ & 32.16 $s$\\
 & 3rd object &  0.05 $s$  & 8.41 $s$  &  17.28 $s$ & 7.77 $s$ & 13.15 $s$ & 46.66 $s$\\
 \midrule
 \multirow{ 3}{*}{Coke cans} 
 & 1st object & 0.06 $s$  & 13.92 $s$ & 4.95 $s$ &  8.21 $s$ & 3.67 $s$ & 30.81 $s$\\
 & 2nd object & 0.06 $s$  & 7.98 $s$  & 11.65 $s$ & 9.12 $s$ & 13.33 $s$ & 42.14 $s$\\
\midrule
 \multirow{ 4}{*}{Tissue boxes} 
 & 1st object & 0.07 $s$  & 13.81 $s$ & 4.79 $s$ & 7.97 $s$ & 10.71 $s$ & 37.35 $s$\\
 & 2nd object & 0.06 $s$  & 8.49 $s$  & 12.38 $s$ & 8.46 $s$ & 4.86 $s$ & 34.25 $s$\\
 & 3rd object & 0.07 $s$  & 7.74 $s$  & 16.33 $s$ & 8.60 $s$ & 5.30 $s$ & 38.04 $s$\\ 
\midrule
\bottomrule
\end{tabular}
 \begin{tablenotes}
        The values are based on the average results of ten simulations.
      \end{tablenotes}
\end{threeparttable}
}

\end{table*}

The computer configuration used for doing the experiment is Intel Core i9-9900K CPU, 32 GB memory with 1600 MHz and GeForce GTX 1080Ti GPU. The programming language used for developing the system is Python.
Table \ref{tb:timeconsuming} is the time consumption in the simulation.
The unit is second. In the table, the order of the objects is determined by their Euclidean distances the goal position.
The average total time for moving the tissue boxes is about 110 seconds for ten times of simulation.
The average total time for moving the Domino blocks is about 116 seconds for ten times of simulation.
The average total time of piling two coke cans up is around 73 seconds for ten times of simulation.
Although only suction poses on the top facet is considered, the planner successfully finds feasible motion
for all the tests, which indicates the generality and robustness of the proposed combined task and motion planner.

In detail, the suction pose sub-planner takes little time and does not occupy the total time of planning.
The first regrasp component of the first object takes almost twice as much time as the others,
because it needs to add the grasps of both the tool at the initial pose and the suction poses on the object.
The other regrasp component uses the final grasp of the tool in the previous planning as the initial grasp
and only needs to add the grasps at the final goal pose.
The time consumption of regrasp component takes around 50 percent of total time. The time spent 
by the motion component is very variable. This is because the recognized 
object pose may have a 180 degrees rotation difference around the vertical direction. 
Robots may need to do a 180-degree rotation motion, which increases the time of the motion component.

\begin{figure}[!htbp]
    \centerline{\includegraphics[width=\linewidth]{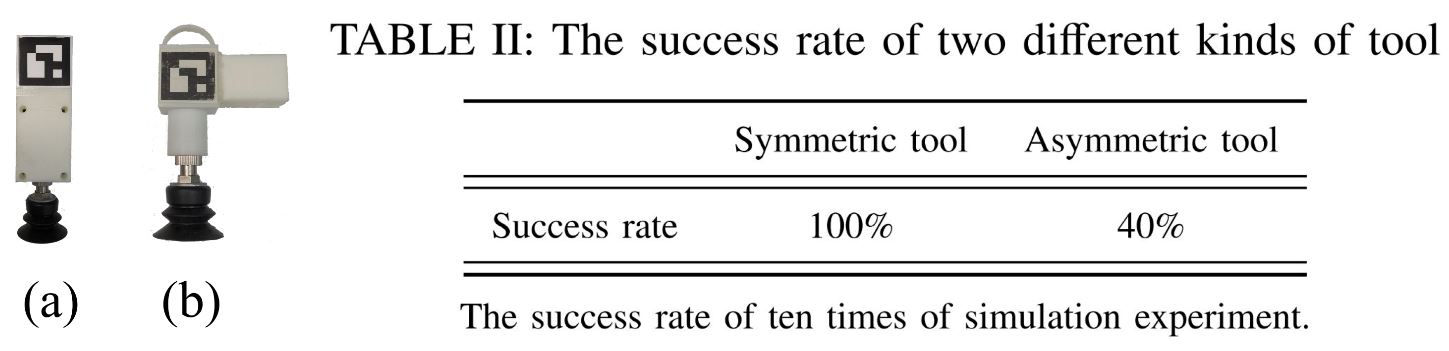}}
    \caption{(a) A long and symmetric suction up tool. (b) A short and asymmetric suction cup tool.
    Table II: The success rate of the two different tools.}
    \label{toolcompare}
\end{figure}

We also compared different designs of the suction cup tool. 
Fig.\ref{toolcompare}(a) is the design used in Fig.\ref{simulation}. The structure of the tool is long and symmetric. 
It has a large area for left hand and right hand to do the handover and
is easy to adjust the grasp pose by the handover.
We compared it with another tool shown in Fig.\ref{toolcompare}(b). 
Compared with the first one, the second tool is short and asymmetric.
There is a handle for helping handover. However, the space for the grasp is not enough and
the effective handover grasp is much less than the symmetric tool.
The time of planning increases a lot because of the design.
The success rate is also affected by the tool. The Table II embedded in Fig.\ref{toolcompare}
compares the success rate of using the two different tools for moving the three tissue boxes ten times
in the simulation. The success rate of using the long and symmetric tool is much higher than using the short and asymmetric tool.
It indicates that it is better to design the symmetrical structure tool for the proposed combined task
and motion planner, which can increase the number of handover and the success rate.
In addition, the tool with more effective grasps can lead to better results as well.

\subsection{Real-world executions}
\begin{figure*}[!htbp]
\centerline{\includegraphics[width=.97\textwidth]{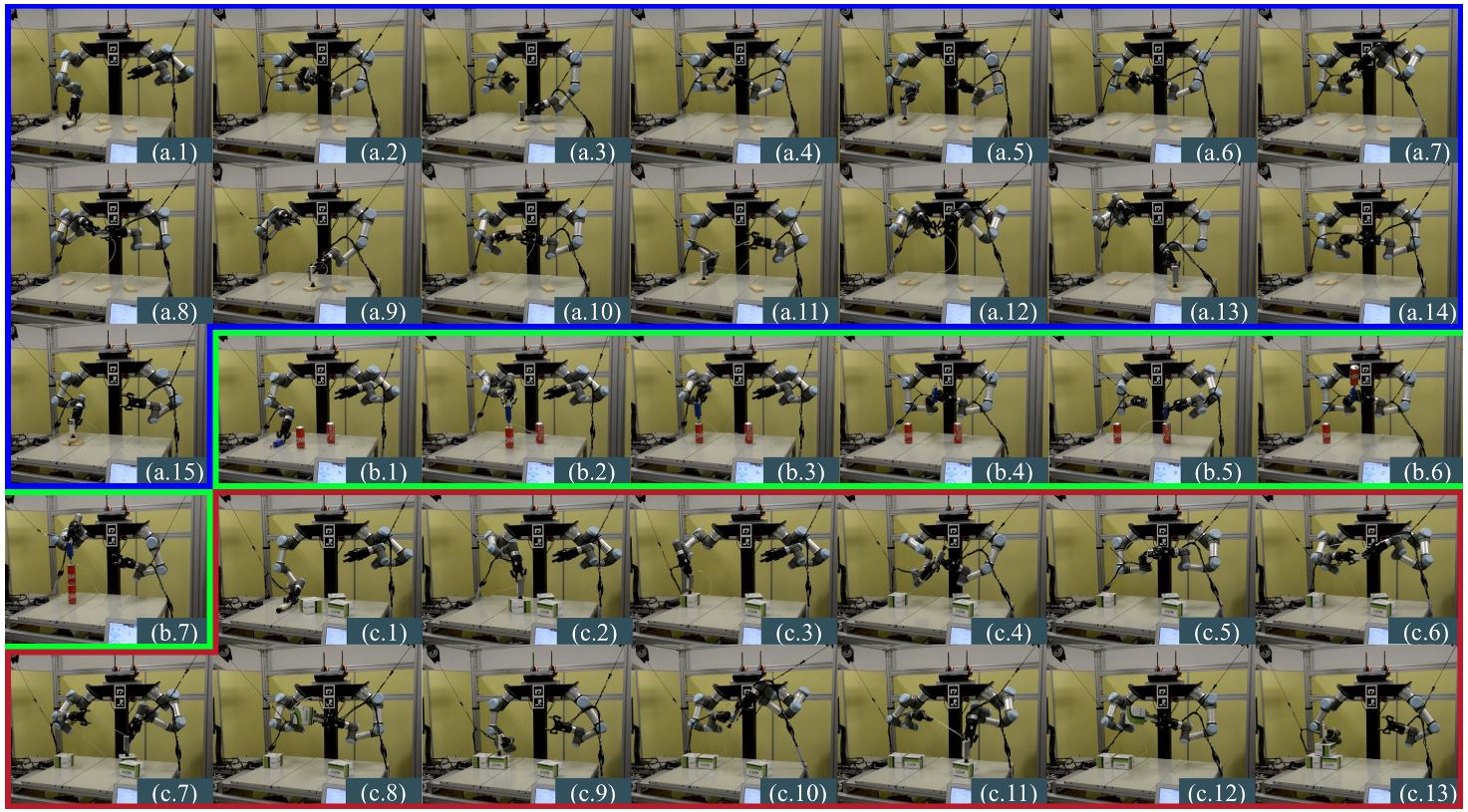}}
\caption{The real-world executions of previous simulation results.
(a) The robot moves three Domino blocks to the goal. (b) The robot piles up two Coke cans.
(c) The robot moves three tissue boxes to the goal.}
\label{realworldexecutedomino}
\end{figure*}

Fig.\ref{realworldexecutedomino} shows the results of real-world executions.
We used the tool with a big suction pad for the Domino blocks and tissue boxes,
and used the tool with a small suction cup for Coke cans.

One problem we found in the real-world execution is we did not consider the vacuum pipe.
Sometimes the pipe wound around the robot arm, which might change the pose of the object or the suction cup tool 
and lead to failure. We are developing algorithms to solve this problem in another work \cite{daniel2019}
and will integrate them
into this system at a later stage.

\section{Conclusions and Future Work}
The paper proposed a combined task and motion planner for the robot to use a suction cup tool to move
objects. The planner enables a robot to use tools to perform tasks that are beyond
the generality of a classical robotic gripper. The proposed planner is implemented and validated using a
dual-arm robot system. The experimental results show that the planner is efficient and robust.
Our future work will be including force control in the planning loop.


\bibliographystyle{IEEEtran}
\bibliography{reference.bib}

\begin{thebibliography}{10}
\providecommand{\url}[1]{#1}
\csname url@samestyle\endcsname
\providecommand{\newblock}{\relax}
\providecommand{\bibinfo}[2]{#2}
\providecommand{\BIBentrySTDinterwordspacing}{\spaceskip=0pt\relax}
\providecommand{\BIBentryALTinterwordstretchfactor}{4}
\providecommand{\BIBentryALTinterwordspacing}{\spaceskip=\fontdimen2\font plus
\BIBentryALTinterwordstretchfactor\fontdimen3\font minus
  \fontdimen4\font\relax}
\providecommand{\BIBforeignlanguage}[2]{{%
\expandafter\ifx\csname l@#1\endcsname\relax
\typeout{** WARNING: IEEEtran.bst: No hyphenation pattern has been}%
\typeout{** loaded for the language `#1'. Using the pattern for}%
\typeout{** the default language instead.}%
\else
\language=\csname l@#1\endcsname
\fi
#2}}
\providecommand{\BIBdecl}{\relax}
\BIBdecl

\bibitem{bibhasegawa2017three}
S.~Hasegawa, K.~Wada, Y.~Niitani, K.~Okada, and M.~Inaba, ``A three-fingered
  hand with a suction gripping system for picking various objects in cluttered
  narrow space,'' in \emph{IEEE International Conference on Intelligent Robots
  and Systems (IROS)}, 2017, pp. 1164--1171.

\bibitem{takahashi2013flexible}
T.~Takahashi, K.~Nagato, M.~Suzuki, and S.~Aoyagi, ``Flexible vacuum gripper
  with autonomous switchable valves,'' in \emph{IEEE International Conference
  on Robotics and Automation (ICRA)}, 2013, pp. 364--369.

\bibitem{yamaguchi2013development}
K.~Yamaguchi, Y.~Hirata, and K.~Kosuge, ``Development of robot hand with
  suction mechanism for robust and dexterous grasping,'' in \emph{IEEE
  International Conference on Intelligent Robots and Systems (IROS)}, 2013, pp.
  5500--5505.

\bibitem{hu2019designing}
Z.~Hu, W.~Wan, and K.~Harada, ``Designing a mechanical tool for robots with
  2-finger parallel grippers,'' \emph{IEEE Robotics and Automation Letters},
  vol.~4, pp. 2981--2988, 2019.

\bibitem{cambon2009hybrid}
S.~Cambon, R.~Alami, and F.~Gravot, ``A hybrid approach to intricate motion,
  manipulation and task planning,'' \emph{The International Journal of Robotics
  Research}, vol.~28, no.~1, pp. 104--126, 2009.

\bibitem{wolfe2010combined}
J.~Wolfe, B.~Marthi, and S.~Russell, ``Combined task and motion planning for
  mobile manipulation,'' in \emph{International Conference on Automated
  Planning and Scheduling (ICAPS)}, 2010.

\bibitem{kaelbling2010hierarchical}
L.~P. Kaelbling and T.~Lozano-P{\'e}rez, ``Hierarchical planning in the now,''
  in \emph{Workshops at the Twenty-Fourth AAAI Conference on Artificial
  Intelligence}, 2010.

\bibitem{faser2016}
L.~{Fraser}, B.~{Rekabdar}, M.~{Nicolescu}, M.~{Nicolescu}, D.~{Feil-Seifer},
  and G.~{Bebis}, ``A compact task representation for hierarchical robot
  control,'' in \emph{IEEE-RAS International Conference on Humanoid Robots
  (Humanoids)}, 2016, pp. 697--704.

\bibitem{bidot2017geometric}
J.~Bidot, L.~Karlsson, F.~Lagriffoul, and A.~Saffiotti, ``Geometric
  backtracking for combined task and motion planning in robotic systems,''
  \emph{Artificial Intelligence}, vol. 247, pp. 229--265, 2017.

\bibitem{suarez2018interleaving}
A.~Su{\'a}rez-Hern{\'a}ndez, G.~Aleny{\`a}, and C.~Torras, ``Interleaving
  hierarchical task planning and motion constraint testing for dual-arm
  manipulation,'' in \emph{2018 IEEE International Conference on Intelligent
  Robots and Systems (IROS)}, 2018, pp. 4061--4066.

\bibitem{colledanchise2019}
M.~Collendanchise, D.~Almeida, and P.~Ogren, ``Towards blended reactive
  planning and acting using behavior trees,'' in \emph{IEEE International
  Conference on Robotics and Automation (ICRA)}, 2000.

\bibitem{harada2014manipulation}
K.~Harada, T.~Tsuji, and J.-P. Laumond, ``A manipulation motion planner for
  dual-arm industrial manipulators,'' in \emph{IEEE International Conference on
  Robotics and Automation (ICRA)}, 2014, pp. 928--934.

\bibitem{suarez2015using}
R.~Su{\'a}rez, J.~Rosell, and N.~Garcia, ``Using synergies in dual-arm
  manipulation tasks,'' in \emph{IEEE International Conference on Robotics and
  Automation (ICRA)}, 2015, pp. 5655--5661.

\bibitem{moriyama2019dual}
R.~Moriyama, W.~Wan, and K.~Harada, ``Dual-arm assembly planning considering
  gravitational constraints,'' 2019.

\bibitem{wan2016achieving}
W.~{Wan} and K.~{Harada}, ``Achieving high success rate in dual-arm handover
  using large number of candidate grasps, handover heuristics, and hierarchical
  search,'' \emph{Advanced Robotics}, vol.~30, no. 17-18, pp. 1111--1125, 2016.

\bibitem{wan2019preparatory}
W.~{Wan}, K.~{Harada}, and K.~{Fumio}, ``Preparatory manipulation planning
  using automatically determined single and dual arms,'' \emph{IEEE
  Transactions on Industrial Informatics}, 2019.

\bibitem{kuffner2000rrt}
J.~J. Kuffner~Jr and S.~M. LaValle, ``Rrt-connect: An efficient approach to
  single-query path planning,'' in \emph{IEEE International Conference on
  Robotics and Automation (ICRA)}, 2019.

\bibitem{daniel2019}
D.~Sanchez, W.~Wan, and K.~Harada, ``Arm manipulation planning of tethered
  tools with the help of a tool balancer,'' in \emph{IFToMM World Congress},
  2019.

\end{thebibliography}

\end{document}